\documentclass[11pt]{article}

\usepackage[preprint]{acl}

\usepackage{times}
\usepackage{booktabs}
\usepackage{latexsym}
\usepackage{subcaption}
\usepackage[T1]{fontenc}
\usepackage{float}
\usepackage[utf8]{inputenc}
\usepackage[T1]{fontenc}    % use 8-bit T1 fonts
\usepackage{hyperref}       % hyperlinks
\usepackage{url}            % simple URL typesetting
\usepackage{booktabs}       % professional-quality tables
\usepackage{threeparttable}
\usepackage{amsfonts}       % blackboard math symbols
\usepackage{nicefrac}       % compact symbols for 1/2, etc.
\usepackage{microtype}      % microtypography
\usepackage{xcolor}         % colors
\usepackage{amsthm}                % For theorem environments
\usepackage{algorithm}      % 算法浮动体环境
\usepackage{algpseudocode}  % 用于书写伪代码
\usepackage{tabularx}       % 提供 tabularx 环境，可设置表格总宽度
\newcolumntype{C}{>{\centering\arraybackslash}X}

\usepackage{amsmath}
\usepackage{booktabs}   % 用于 \toprule, \midrule, \bottomrule 等三线表命令
\usepackage{siunitx}    % 用于 S 列，实现小数点对齐
\usepackage{multirow}   % 用于 \multirow，实现单元格跨行
\usepackage{caption}    % 推荐使用，可以更好地控制图表标题
\usepackage{microtype}

\usepackage{tcolorbox}
\definecolor{darkorange}{RGB}{255, 140, 0}
\definecolor{lightgreen}{RGB}{145, 204, 117}
\definecolor{gray}{RGB}{128, 128, 128}
\definecolor{lightyellow}{RGB}{250, 200, 88}
\definecolor{lightred}{RGB}{238, 102, 102}
\definecolor{lightblue}{RGB}{115, 192, 222}
\newtcolorbox{promptbox}[2][Prompt]{
colback=black!5!white,
arc=5pt, 
boxrule=0.5pt,
fonttitle=\bfseries,
title=#1, 
before upper={\scriptsize}, fontupper=\fontfamily{ptm}\selectfont,
colframe=#2, % 使用传递的参数来设定 colframe
}
\usepackage{inconsolata}

\usepackage{graphicx}
\usepackage{enumitem}

\title{MaP: A Unified Framework \\ for Reliable Evaluation of Pre-training Dynamics}
\author{
    \textbf{
        Jiapeng Wang\textsuperscript{\rm{1}\thanks{\ \ Equal contribution.}},
        Changxin Tian\textsuperscript{\rm{2}\footnotemark[1]},
        Kunlong Chen\textsuperscript{\rm{2}\ },
        Ziqi Liu\textsuperscript{\rm{2}\ },
    } \\
    \textbf{
        Jiaxin Mao\textsuperscript{\rm{1}\ },
        Wayne Xin Zhao\textsuperscript{\rm{1}\thanks{\ \ Corresponding author.}},
        Zhiqiang Zhang\textsuperscript{\rm{2}\footnotemark[2] },
        Jun Zhou\textsuperscript{\rm{2}\ },
    } \\
    \textsuperscript{1}Gaoling School of Artificial Intelligence, Renmin University of China 
    \textsuperscript{2}Ant Group \\
    \texttt{wangjp1010@ruc.edu.cn, batmanfly@gmail.com}
}

\begin{document}
\maketitle
\begin{abstract}
Reliable evaluation is fundamental to the progress of Large Language Models (LLMs), yet the evaluation process during pre-training is plagued by significant instability that obscures true learning dynamics. 
In this work, we systematically diagnose this instability, attributing it to two distinct sources: \textit{Parameter Instability} from training stochasticity and \textit{Evaluation Instability} from noisy measurement protocols. 
To counteract both sources of noise, we introduce \textbf{MaP}, a dual-pronged framework that synergistically integrates checkpoint \underline{M}erging \underline{a}nd the \underline{P}ass@k metric. Checkpoint merging smooths the parameter space by averaging recent model weights, while Pass@k provides a robust, low-variance statistical estimate of model capability. 
Extensive experiments show that MaP yields significantly smoother performance curves, reduces inter-run variance, and ensures more consistent model rankings. 
Ultimately, MaP provides a more reliable and faithful lens for observing LLM training dynamics, laying a crucial empirical foundation for LLM research.

\end{abstract}

\section{Introduction}
% Large Language Models (LLMs) have demonstrated remarkable capabilities across a broad spectrum of domains — from complex reasoning and domain-specific expertise to creative generation and software engineering — fundamentally reshaping the landscape of artificial intelligence~\citep{zhao2023survey,naveed2025comprehensive}. This rapid progress is fueled by a tightly coupled cycle of model development and evaluation, wherein benchmarks and public leaderboards serve as essential feedback mechanisms~\citep{cao2025toward}. They enable researchers to measure model capabilities, compare methodologies, validate innovations, and steer future directions. 
Large Language Models (LLMs) have demonstrated remarkable capabilities across diverse domains, fundamentally reshaping the artificial intelligence landscape~\citep{zhao2023survey}. This rapid progress is driven by a tightly coupled cycle of development and evaluation, where benchmarks and public leaderboards serve as essential feedback mechanisms~\citep{cao2025toward}. They enable researchers to measure model capabilities, compare methodologies, and guide future research. Consequently, considerable effort has focused on designing benchmarks to evaluate what capabilities a model possesses~\citep{ni2025survey}.

\begin{figure*}[t!]
    \centering
    \begin{subfigure}[b]{0.315\textwidth}
        \centering
        \includegraphics[width=\linewidth]{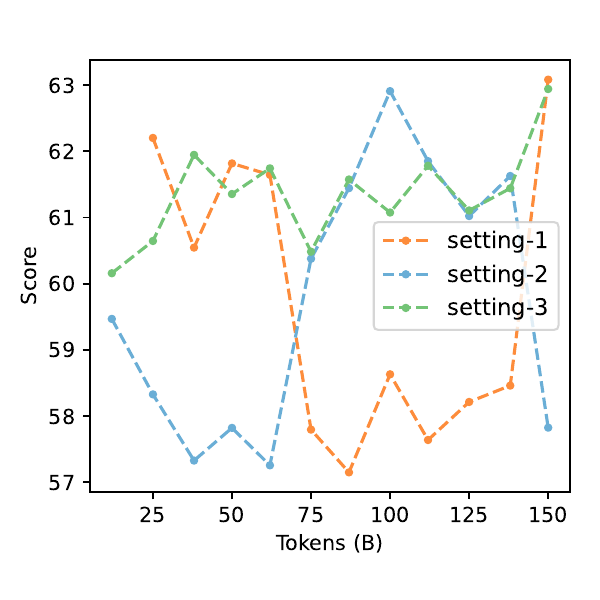}
        \caption{Ambiguous ablation results}
        \label{fig:sub_a}
    \end{subfigure}%
    \hfill
    \begin{subfigure}[b]{0.315\textwidth}
        \centering
        \includegraphics[width=\linewidth]{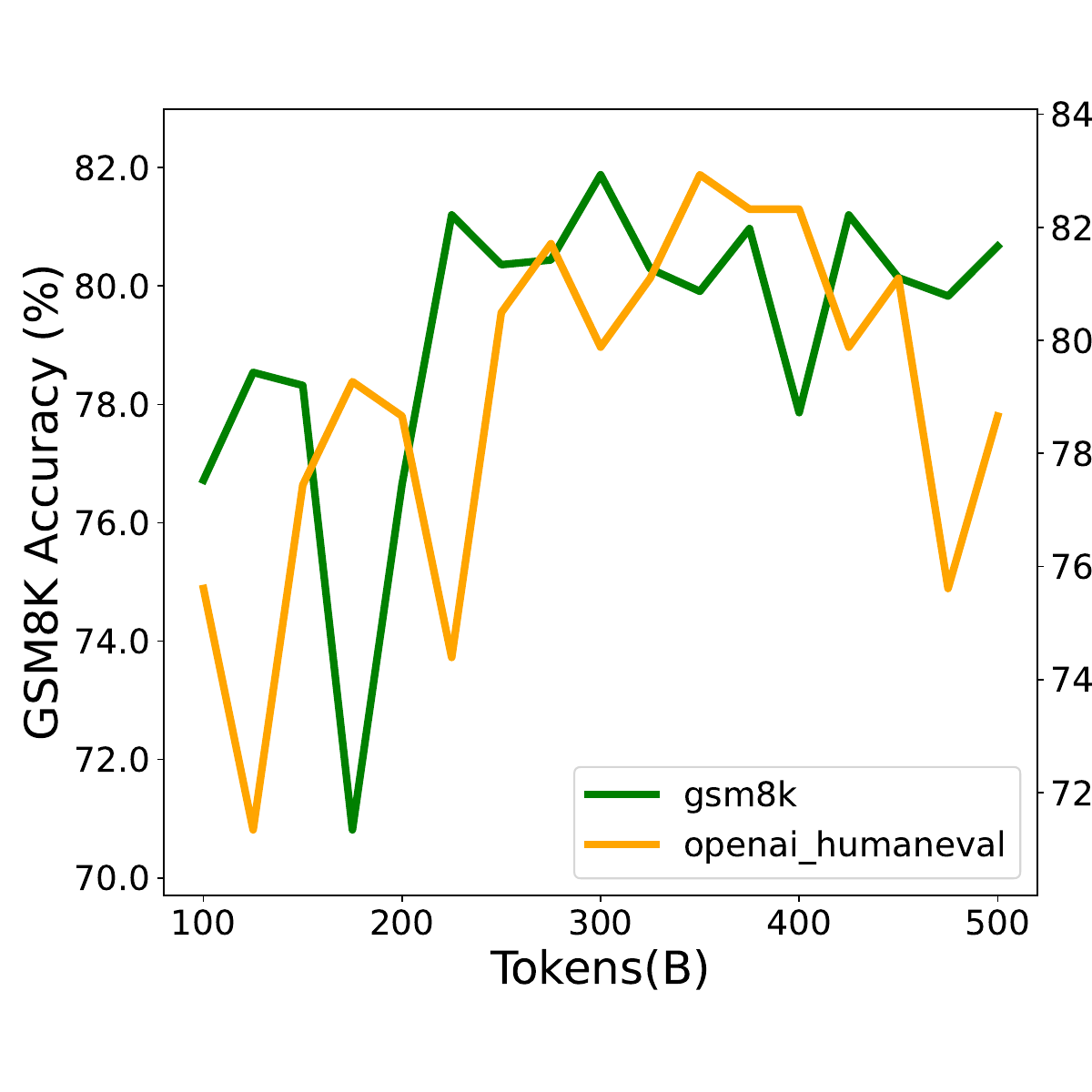}
        \caption{Unstable performance trajectories}
        \label{fig:sub_b}
    \end{subfigure}%
    \hfill
    \begin{subfigure}[b]{0.315\textwidth}
        \centering
        \includegraphics[width=\linewidth]{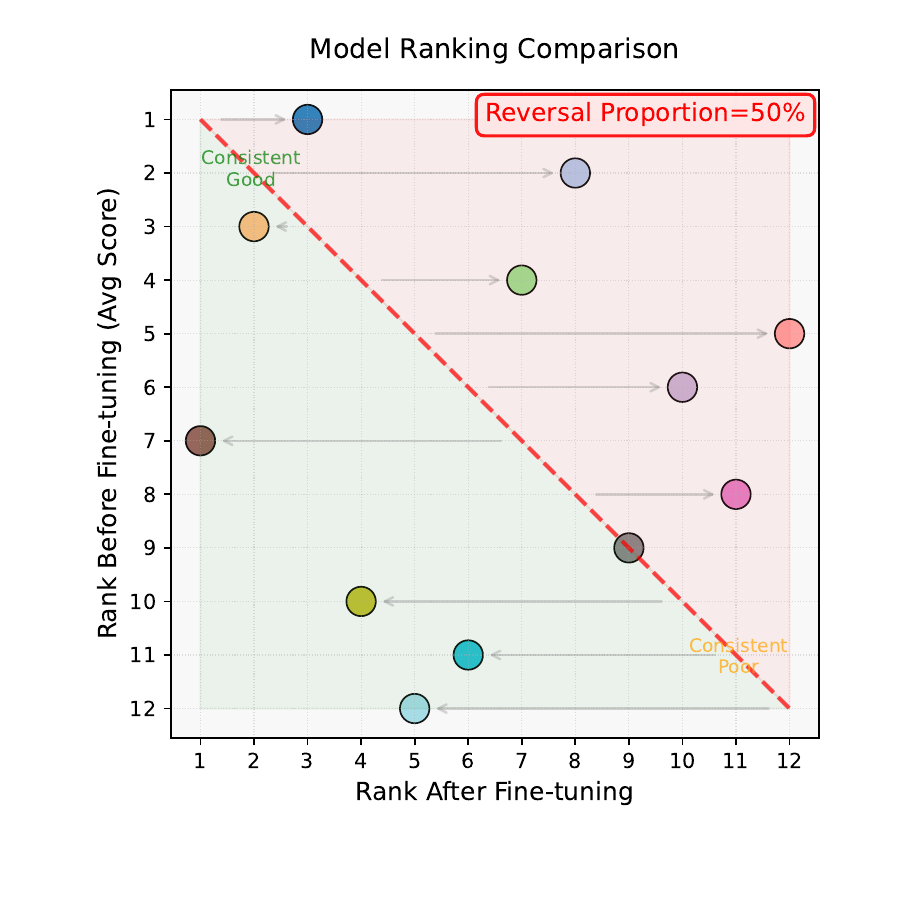}
        \caption{Pre- vs.\ post-training inconsistency}
        \label{fig:sub_c}
    \end{subfigure}%
    \caption{
    Illustrations of evaluation instability during pre-training.
    (a) When comparing training strategies, performance curves often intersect, obscuring which strategy is truly superior. 
    (b) The performance of a single model can be highly volatile during pre-training, which may conceal underlying issues with the training process.
    (c) A rank correlation analysis shows a severe mismatch between the rankings of pre-trained models and their fine-tuned counterparts, indicating that pre-training evaluation often fails to reliably predict final downstream performance.}
    \label{fig:intro}
\end{figure*}

% Yet, while considerable effort has been devoted to designing benchmarks that probe \textit{what} capabilities a model possesses, far less attention has been paid to the procedural rigor of \textit{how} these capabilities are measured—specifically, how to obtain a stable, reproducible, and statistically sound signal to observe the learning dynamics during the expensive pre-training phase~\citep{rybakov2024methods}. In reality, this oversight is significant. Beneath the surface of benchmark leaderboards and scores lies a critical vulnerability: \textit{the evaluation process itself is often unstable and inconsistent}~\citep{lunardi2025robustness}. As illustrated in Figure~\ref{fig:intro}, this instability manifests in multiple concerning ways — volatility in performance trajectories, ambiguous comparisons between training strategies, and poor rank correlation between pretraining and downstream performance. These phenomena obscure genuine progress during training, invalidate ablation studies, and hinder the community’s ability to achieve consistent, reproducible improvements — collectively undermining the validity of our scientific conclusions.
However, the reliability of these evaluations is challenged by the inherent volatility of pre-training, where drastic parameter updates and fragile instruction-following abilities introduce significant noise. As illustrated in Figure~\ref{fig:intro}, this evaluation instability critically impacts the entire LLM pre-training lifecycle: 
(1) \textbf{In ablations studies,} it leads to ambiguous comparisons and can even yield incorrect conclusions (Figure~\ref{fig:sub_a}). 
(2) \textbf{During the pre-training process,} it causes volatile performance trajectories that obscure underlying data or training issues (Figure~\ref{fig:sub_b}). 
(3) \textbf{After pre-training,} it results in a poor correlation between the observed base model's performance and its downstream counterparts, meaning improvements do not reliably transfer to post-training tasks (Figure~\ref{fig:sub_c}). % ~\citep{lunardi2025robustness}.
These phenomena obscure genuine training progress and invalidate ablation studies, ultimately undermining the validity of our scientific conclusions. 
This raises a critical question: 
\begin{center}
\emph{how to reliably measure learning dynamics \\during LLM pre-training?}
\end{center}

\captionsetup[subfigure]{skip=2pt}

To address this issue, we systematically diagnose evaluation instability, tracing it to two primary and compounding sources:
(1) \textbf{Parameter Instability}: This variance arises from the stochastic nature of the training trajectory. An LLM’s path through the high-dimensional parameter space is inherently non-smooth, meaning any single checkpoint may represent a transiently suboptimal state or a sharp local minimum~\citep{jastrzkebski2017three,hansen1992testing,fort2020deep}. Consequently, performance can fluctuate dramatically, rendering evaluations at any single checkpoint an unreliable and noisy snapshot of the model’s true capability at that training stage~\citep{achille2017critical}. 
(2) \textbf{Evaluation Instability}: This variance is introduced by the fragility of measurement protocols. 
For generative tasks such as code generation or mathematical reasoning, metrics based on a single output (e.g., greedy decoding) resemble high-variance Bernoulli trials. The resulting scores are highly sensitive to sampling ``luck'', where a small number of favorable or unfavorable generations can disproportionately affect evaluation outcomes~\citep{song2024good,xia2025evaluating,mirzadeh2024gsm}. 

% Based on this diagnosis, and in contrast to conventional methods that evaluate a single checkpoint with a single output, we propose a principled, dual-pronged framework that applies targeted interventions to each source of noise. 
Based on this diagnosis, we propose \textbf{MaP}, a dual-pronged framework that leverages checkpoint \underline{M}erging \underline{a}nd the \underline{P}ass@k metric to handle both sources of noise. 
First, to mitigate parameter instability, we employ Checkpoint Merging~\citep{Averaging,Modelsoups}. By averaging the weights of recent checkpoints, this technique smooths the training trajectory in parameter space, yielding a more stable estimate of the model's central tendency.
Second, to address evaluation instability in generative reasoning tasks, we adopt the Pass@k metric~\citep{humaneval}. This replaces high-variance, single-output measurements with a robust, low-variance statistical estimate of the model's success probability, hardening the evaluation against sampling randomness. 
While these two techniques have been explored in prior work, they have largely been used in isolation and for distinct purposes. Specifically, checkpoint merging has primarily been applied to enhance final model performance~\citep{commanda,WARP,pma,wsm}, whereas Pass@k serves as a standard for code evaluation or for probing a model's latent potential~\citep{humaneval,mbpp,yue2025does}. 
In contrast, we are the first to synergistically integrate these methods into a unified framework, specifically designed to stabilize pre-training evaluation and offer a more reliable lens through which to observe training dynamics.

% Our extensive experiments across a diverse suite of benchmarks provide compelling evidence that our combined framework yields demonstrably smoother performance trajectories, significantly reduces inter-run variance, and ensures more consistent model rankings while mitigating the “ranking reversal” phenomenon. 
Our extensive experiments across a diverse suite of benchmarks demonstrate that MaP yields smoother performance trajectories, reduces inter-run variance, and ensures more consistent model rankings. 
Crucially, this heightened stability provides a significantly clearer and more faithful signal of model progress during the pre-training phase.
In summary, our core contributions are as follows:

(1) We identify and systematically diagnose the long-standing yet under-recognized problem of evaluation instability in pre-training, decoupling it into the distinct sources of \textit{parameter instability} and \textit{evaluation instability}.

(2) We introduce MaP, a simple yet effective framework that synergistically integrates checkpoint merging and Pass@k address both sources of instability, thereby establishing a robust and stable evaluation pipeline for pre-training.

(3) Through comprehensive experiments, we demonstrate that MaP enables a more reliable observation of training dynamics, effectively mitigating misleading artifacts and laying an empirical foundation for the LLM research community.

\section{Method}

 % we propose a dual-pronged framework designed to systematically suppress variance from its two primary sources. Our approach first stabilizes the model's parameters via \textbf{Model Merging} to counteract the stochasticity of training, and then robustifies the measurement process via \textbf{Pass@k} to mitigate the fragility of evaluation.

\subsection{Conceptual Framework}
% Motivated by the instability illustrated in Figure~\ref{fig:intro}, we posit that the overall stability of the observed LLM performance can be decomposed into two orthogonal components: \textit{parameter stability} and \textit{evaluation stability}. Conceptually, this relationship can be approximated as:
To formalize the instability observed in Figure~\ref{fig:intro}, we decompose the overall stability of LLM performance into two distinct factors: \textit{parameter stability} and \textit{evaluation stability}. 
Conceptually, this relationship can be approximated as:
\begin{equation*}
\begin{split}
    & \textit{Overall Stability} \\
    \approx{} & \textit{Parameter Stability} \times \textit{Evaluation Stability}.
\end{split}
\end{equation*}
Our framework aims to maximize overall stability by independently optimizing both components.

\paragraph{Parameter Stability}reflects the extent to which a checkpoint’s performance at a given training stage is representative of its true capability, rather than an artifact of transient parameter configurations. Due to optimization stochasticity, individual checkpoints may occupy noisy or atypical regions of the loss landscape, leading to erratic performance fluctuations. When parameter instability is high, checkpoint-based evaluations become unreliable indicators of learning progress, masking the model’s underlying trajectory. 

\paragraph{Evaluation Stability} refers to the consistency of performance scores produced by the evaluation protocol for a \textit{fixed} set of model parameters. Instability in this component is primarily driven by measurement noise — for instance, randomness in generative sampling.

\subsection{Mitigating Parameter Instability via Checkpoint Merging}
\label{sec:method_merge}
The training trajectory of an LLM is a noisy path through a high-dimensional parameter space. A single checkpoint may represent a transiently suboptimal state or a sharp local minimum, rather than the model’s robust, generalized capability~\citep{hansen1992testing}.
However, previous work predominantly relies on evaluating a single checkpoint, which makes the reported performance highly sensitive to these random fluctuations and yields an unreliable measure of the model's true potential.
While weight averaging is known to construct versatile models~\citep{commanda,WARP} or improve generalization~\citep{sanyal2023early,Checkpoint,lawa,Trainable}, we repurpose it as a dynamic stabilization tool to denoise model parameters at each evaluation step. We implement this as a non-intrusive, post-hoc evaluation protocol that is strictly decoupled from the optimization loop. Specifically, let $\theta_t$ be the parameter vector at training step $t$. At each evaluation step $T$, we define a merging window of size $N$ and compute the element-wise average of the recent $N$ checkpoints saved by the training process to form a surrogate model $\hat{\theta}_T$:
\begin{equation}
    \hat{\theta}_T = \frac{1}{N} \sum_{i=0}^{N-1} \theta_{T-i}.
\end{equation}
This surrogate model $\hat{\theta}_T$ is used \textit{solely} for evaluation; the actual training continues seamlessly from the latest raw checkpoint $\theta_T$, ensuring that the training dynamics remain unaltered. This procedure smooths out high-frequency fluctuations in the parameter space, effectively finding a more stable central point in the recent training trajectory. This merged model is hypothesized to reside in a broader, flatter region of the loss landscape, leading to more robust and representative performance.

\paragraph{A Statistical View of Parameter Stabilization}
Let $\theta_t^*$ denote the ideal parameter vector that captures the central tendency of the training process at step $t$. A checkpoint saved at step $t$ can be modeled as this ideal vector corrupted by zero-mean noise, $\epsilon_t$, arising from stochastic factors such as data batching and dropout:
\begin{equation*}
    \theta_t = \theta_t^* + \epsilon_t, \quad \text{where } E[\epsilon_t] = 0.
\end{equation*}
The performance of a single checkpoint $\theta_t$ is thus a noisy representation of the performance of $\theta_t^*$. If we assume the noise vectors from recent, nearby checkpoints $\{\epsilon_{T-N+1}, \dots, \epsilon_T\}$ are approximately independent, we can analyze the effect of averaging. The merged model $\hat{\theta}_T$ is:

\begin{equation*}
\begin{split}
    \hat{\theta}_T & = \frac{1}{N} \sum_{i=0}^{N-1} \theta_{T-i} = \frac{1}{N} \sum_{i=0}^{N-1} (\theta_{T-i}^* + \epsilon_{T-i}) \\
    & = \frac{1}{N} \sum_{i=0}^{N-1} \theta_{T-i}^* + \frac{1}{N} \sum_{i=0}^{N-1} \epsilon_{T-i}
\end{split}
\end{equation*}

\(\frac{1}{N} \sum_{i=0}^{N-1} \theta_{T-i}^* \) approximates the ideal model obtained by applying learning rate annealing along the ideal training trajectory~\cite{deepseekv3,wsm}.
If $\text{Var}(\epsilon_t) = \Sigma$, then the variance of the averaged noise is:
\begin{equation*}
    \text{Var}\left(\frac{1}{N} \sum_{i=0}^{N-1} \epsilon_{T-i}\right) = \frac{1}{N^2} \sum_{i=0}^{N-1} \text{Var}(\epsilon_{T-i}) = \frac{\Sigma}{N}
\end{equation*}
Thus, checkpoint merging reduces the variance of the parameter noise by a factor of $N$. The resulting model $\hat{\theta}_T$ is a statistically more stable estimate of the ideal model $\theta_T^*$, providing a significantly more reliable basis for evaluation.
A more detailed discussion on the impact of gradient correlation and the validity of this approximation are provided in Appendix~\ref{app:variance_analysis}.
\subsection{Mitigating Evaluation Instability via Pass@k}
\label{sec:passk}
For open-ended generative tasks like code and math, metrics such as accuracy on a single greedy-decoded sample are equivalent to a high-variance Bernoulli trial for each problem. The final score is thus highly susceptible to sampling luck.
While the Pass@k protocol \citep{humaneval} has typically been employed in specific domains like code generation~\citep{humaneval,mbpp} or probe the upper bounds of a model's capability~\citep{yue2025does}, we reposition it here through the lens of evaluation stability and consistency.
Our framework elevates Pass@k from a task-specific metric to a general principle for robust generative evaluation. By generating multiple candidate solutions, it moves beyond a simple average over $n$ samples (which is an estimator for Pass@1). Instead, by measuring the probability of generating at least one correct answer in $k$ attempts (where $k > 1$), it employs a less stringent success criterion. This approach has a distinct advantage when assessing downstream consistency, as it better reflects a model's latent potential rather than its performance on a single, high-variance greedy decode. As we will demonstrate in Section \ref{passk}, this makes the resulting evaluation not only more robust by reducing score variance but also a more reliable predictor.
For Multiple-Choice (MC) tasks (e.g., MMLU), we retain the standard greedy or likelihood-based evaluation, as the finite answer space makes multi-sample metrics susceptible to random guessing.
% Following the methodology introduced by \cite{humaneval}, Pass@k can be computed with an unbiased estimator. For each problem, we generate $n$ candidate solutions, where $n \ge k$. If $c$ of these $n$ solutions are correct, we can estimate the probability of generating at least one correct solution in $k$ independent draws. This is equivalent to one minus the probability of all $k$ draws being incorrect. The number of ways to draw $k$ incorrect solutions from the $n-c$ available is $\binom{n-c}{k}$, and the total number of ways to draw any $k$ solutions from $n$ is $\binom{n}{k}$. Thus, the unbiased estimator is:
% \begin{equation*}
%     \text{pass}@k = 1 - \frac{\binom{n-c}{k}}{\binom{n}{k}}
% \end{equation*}
\paragraph{A Probabilistic View of Measurement Stabilization}
% For a given problem, let $p$ denote the model's latent probability of generating a correct solution in a single attempt. A standard evaluation observes a single outcome, $s \sim \text{Bernoulli}(p)$, which is a high-variance representation of the underlying probability:
% \begin{equation*}
%     \text{Var}(s) = p(1-p)
% \end{equation*}
% For challenging problems where $p$ is not close to 0 or 1, this variance is substantial, making the observed score highly sensitive to sampling luck.

% Instead of observing a single random outcome $X$, we aim to directly estimate the underlying parameter $p$. By generating $n$ samples, we draw from the distribution to form a more robust estimate. The Pass@k formula is an estimator, $\hat{p}_k$, for the probability of at least one success in $k$ trials, which equals $1 - (1-p)^k$. An estimator's quality is measured by its own variance, $\text{Var}(\hat{p}_k)$. By leveraging information from $n$ samples, the variance of this estimator is significantly lower than the variance of the single Bernoulli outcome:
% \begin{equation*}
%     \text{Var}(\hat{p}_k) \ll \text{Var}(X)
% \end{equation*}
% In essence, Pass@k trades a modest increase in computational cost for a massive reduction in measurement error. It replaces a high-variance single data point with a low-variance statistical estimate, ensuring the final score faithfully reflects the model's capability.
For a given problem, let $p$ denote the model's latent probability of generating a correct solution in a single attempt. A standard evaluation observes a single outcome, $s \sim \text{Bernoulli}(p)$. The variance of this single-trial measurement is high:
\begin{equation*}
    \text{Var}(s) = p(1-p)
\end{equation*}
For challenging problems where $p$ is not close to 0 or 1, this variance is substantial, making the observed outcome highly sensitive to sampling luck.

Instead of this noisy single-point measurement, the Pass@k protocol leverages $n$ independent samples to provide a stable, unbiased estimate of the model's capability. As we formally derive in Appendix \ref{sec:appendix_pass_at_k}, the variance of the Pass@k estimator, $\hat{q}_k$, can be shown to be approximately:
\begin{equation*}
    \text{Var}(\hat{q}_k) \approx \frac{k^2(1-p)^{2(k-1)} \cdot p(1-p)}{n}
\end{equation*}
This result demonstrates a substantial reduction in measurement variance. By replacing a high-variance single data point with a low-variance statistical estimate, the final score becomes more stable and consistently reflects the model's true capability.

% takeaway
% *   **Model Merging** improves *parameter stability* by averaging the weights of the last $N$ checkpoints, reducing the parameter noise variance by a factor of **$N$**.
% *   **Pass@k** improves *evaluation stability* by aggregating results from $n$ output samples, reducing measurement variance by a factor of approximately **$n / [k^2(1-p)^{2(k-1)}]$**.

\subsection{The Complete Synergistic Framework}
Our full stabilization framework is the synergistic application of both techniques. At each evaluation point, we first apply Checkpoint Merge to obtain a low-variance model estimate $\hat{\theta}_T$. We then evaluate this stable model using the low-variance Pass@k protocol. By jointly mitigating both noises, we achieve a maximally stable and reliable assessment of model progress.

% \begin{promptbox}[Key Takeaway]{gray}
% Our full stabilization framework jointly mitigating both source noise and measurement noise: \\
% \textbf{1}. Model Merging improves \textbf{parameter stability} by averaging the weights of the last $N$ checkpoints, reducing the parameter noise variance by a factor of $N$.\\
% \textbf{2}. Pass@k improves \textbf{evaluation stability} by aggregating results from $n$ output samples, reducing measurement variance by a factor of approximately $n / [k^2(1-p)^{2(k-1)}]$.
% \end{promptbox}

\begin{center}
\begin{tcolorbox}[
    colback=gray!10, 
    colframe=black!65, 
    width=0.48\textwidth, % 调整宽度
    title={Key Takeaway},
    fonttitle=\normalsize, 
    top=2mm, % 上边距
    bottom=2mm, % 下边距
    left=1mm, % 左边距
    right=1mm % 右边距
]
    \small
    Our full stabilization framework jointly mitigating both source noise and measurement noise:
    \vspace{2mm}
    \begin{itemize}[leftmargin=1em] % 使用 enumitem 宏包调整左边距
        \item 
        \textbf{Checkpoint Merging} improves \textbf{parameter stability} by averaging the weights of the last $N$ checkpoints, theoretically reducing the parameter noise variance by up to a factor of $N$.  
        \item 
        \textbf{Pass@k} improves \textbf{evaluation stability} by aggregating results from $n$ output samples, reducing measurement variance by a factor of $n / [k^2(1-p)^{2(k-1)}]$.
    \end{itemize}
\end{tcolorbox}
\end{center}

\section{Experiment}

\subsection{Experimental Setup}

\paragraph{Metrics for quantifying stability}
To move beyond qualitative observations and quantitatively assess the stability of our evaluation framework, we introduce two specialized metrics.

First, to quantify the stability of the learning trajectory, we measure how consistently performance improves over time. The core intuition is that a stable training process should ensure that a later checkpoint's performance is consistently non-decreasing relative to an earlier one. To formalize this, we compute Kendall's rank correlation coefficient ($\tau$)~\citep{kendall1938new} between the chronological sequence of checkpoints and their evaluation scores~\citep{bose}. A higher $\tau$ indicates a more stable and predictable learning trajectory. It is defined as:
\begin{equation*}
    \tau = \frac{P - (n(n - 1)/2 - P)}{n(n - 1)/2} = \frac{4P}{n(n - 1)} - 1
\end{equation*}
where $n$ is the number of checkpoints, and $P$ is the number of concordant pairs, that is, pairs in which a later checkpoint achieves a higher score than an earlier one. A $\tau$ value of 1 indicates perfect monotonic improvement, while a value near 0 suggests performance fluctuates randomly over time.

Second, to evaluate how well pre-training performance predicts downstream capabilities, we introduce the Pairwise Ranking Reversal Rate (PRR). This metric quantifies the proportion of model pairs whose relative ranking reverses after a subsequent stage such as supervised fine-tuning (SFT). For a set of $M$ models, the PRR is:
\begin{multline*}
    \text{PRR} = \frac{1}{\binom{M}{2}} \sum_{1 \le i < j \le M} \mathbb{I}\Bigl[ (R_{PT}(i) - R_{PT}(j)) \\
    \cdot (R_{SFT}(i) - R_{SFT}(j)) < 0 \Bigr]
\end{multline*}
where $R_{PT}(i)$ and $R_{SFT}(i)$ are the performance ranks of model $i$ before and after fine-tuning, respectively, and $\mathbb{I}[\cdot]$ is the indicator function. A PRR of 0 indicates perfect rank consistency, while a PRR of 0.5 implies that pre-training evaluation provides no predictive signal for downstream rankings.

\paragraph{Models and benchmarks}
To enable fine-grained analysis of the training trajectory, we conduct experiments using intermediate checkpoints from our own pre-trained models. Our model incorporates both dense and Mixture-of-Experts (MoE) architectures, with total parameter counts spanning from 1B to 8B. In the MoE variant, 1.4B parameters are activated. In our analysis experiments, we default to the largest MoE model. Comprehensive details regarding our model architecture and specific training parameters are provided in Appendix~\ref{setup}.

We evaluate performance across various benchmark categories: \textbf{General} (AGIEval~\citep{agieval}, RACE~\citep{race}, SQuAD2.0~\citep{squad2}); \textbf{Knowledge} (e.g., MMLU~\citep{mmlu}, CMMLU~\citep{cmmlu}, C-Eval~\citep{ceval}); \textbf{Math} (e.g., GSM8K~\citep{gsm8k}, MATH~\citep{math}, GSM-Plus~\citep{gsm-plus}, CMATH~\citep{cmath}); and \textbf{Code} (e.g., HumanEval~\citep{humaneval}, MBPP~\citep{mbpp}, HumanEval-Plus~\citep{mbpp+}, MBPP-Plus~\citep{mbpp+}). To manage the computational cost associated with the Pass@k metric, we report its results on a representative subset of key generative tasks: GSM8K, MATH, HumanEval, and MBPP. All experiments are implemented using the OpenCompass framework~\citep{contributors2023opencompass}.

\begin{figure*}[h]
    \centering    \includegraphics[width=1\textwidth]{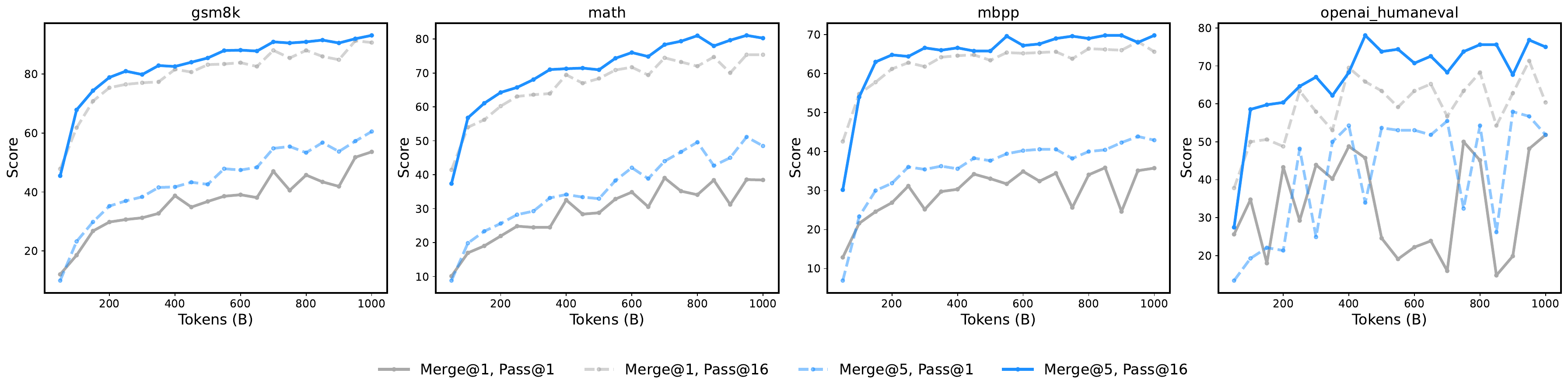}
  \caption{Visual comparison of performance trajectories under different stability protocols.}
  \label{fig:combine}
\end{figure*}

\subsection{The Synergistic Effect of the Unified Framework}
While Checkpoint Merging and Pass@k individually mitigate instability, they target distinct sources of variance. Stabilizing only the model parameters via merging still leaves the evaluation susceptible to measurement noise. Conversely, a robust evaluation protocol like Pass@k cannot compensate for the inherent volatility of the underlying model checkpoints. We therefore hypothesize that their combined application is necessary to achieve maximal stability.
To validate this, we conduct an ablation study during a 1T-token pre-training run. We  evaluate performance using our combined framework and its individual components. The results are presented both qualitatively in Figure~\ref{fig:combine} and quantitatively in Table~\ref{tab:ablation_study}.
Both individual methods show substantial improvement over the baseline protocol, and the full framework achieves the highest or near-highest correlation scores and a visibly smoother trend, demonstrating a clear synergistic effect. 
These findings show that jointly addressing parameter and evaluation instability is crucial for achieving truly reliable LLM evaluation. 

% \begin{table}[th!]
% \centering

% \resizebox{\columnwidth}{!}{%
% \begin{tabular}{lcccc}
% \toprule
% \textbf{Protocol} & \textbf{GSM8K}&\textbf{MATH} & \textbf{HumanEval} & \textbf{MBPP} \\
% \midrule
% Original (Merge@1, Pass@1) &0.863& 0.788 & 0.378 & 0.579 \\
% + Merge@5 & 0.910&0.874 & 0.310 & \textbf{0.839} \\
% + Pass@16 & 0.871&0.850 & 0.374 & 0.794 \\
% \midrule
% \textbf{MaP (Merge@5, Pass@16)} & \textbf{0.926}&\textbf{0.895} & \textbf{0.631} & 0.778 \\
% \bottomrule
% \end{tabular}
% }
% \caption{Ablation study on the combined effects of Model Merging and Pass@k. Merge@$N$ denotes averaging the last $N$ checkpoints. Measured by Kendall's rank correlation ($\tau$).}
% \label{tab:ablation_study}
% \end{table}

\begin{table}[th!]
\centering

% 调整表格宽度适应单栏
\resizebox{\columnwidth}{!}{%
\begin{tabular}{llccccc}
\toprule
\textbf{Model} & \textbf{Protocol} & \textbf{GSM8K} & \textbf{MATH} & \textbf{HumanEval} & \textbf{MBPP} & \textbf{Avg.}\\
\midrule
% --- 16B Section ---
\multirow{4}{*}{\textbf{1B dense}} 
 & Original (M@1, P@1)       & 0.816 & 0.833 & 0.133 & 0.761 &0.635\\
 & + Merge@5                 & 0.812 & 0.800 & \textbf{0.705} & 0.828 &0.786\\
 & + Pass@16                 & 0.800 & 0.889 & 0.366 & 0.783 &0.562\\
 & \textbf{MaP (M@5, P@16)}  & \textbf{0.950} & \textbf{0.891} & 0.492 & \textbf{0.862} & \textbf{0.799}\\

\midrule
% --- 4B Section ---
\multirow{4}{*}{\textbf{4B MoE}} 
 & Original (M@1, P@1)       & 0.801 & 0.663 & 0.274 & 0.824 & 0.640\\
 & + Merge@5                 & 0.826 & 0.701 & 0.369 & \textbf{0.926} & 0.705\\
 & + Pass@16                 & 0.660 & \textbf{0.812} & 0.605 & 0.779 & 0.714\\
 & \textbf{MaP (M@5, P@16)}  & \textbf{0.869} & 0.617 & \textbf{0.750} & 0.878 & \textbf{0.779}\\
\midrule
% --- 8B Section (Original Data) ---
\multirow{4}{*}{\textbf{8B MoE}} 
 & Original (M@1, P@1)       & 0.863 & 0.788 & 0.378 & 0.579 &0.652\\
 & + Merge@5                 & 0.910 & 0.874 & 0.310 & \textbf{0.839} &0.733\\
 & + Pass@16                 & 0.871 & 0.850 & 0.374 & 0.794 &0.722\\
 & \textbf{MaP (M@5, P@16)}  & \textbf{0.926} & \textbf{0.895} & \textbf{0.631} & 0.778 &\textbf{0.808}\\

\bottomrule
\end{tabular}
}
\caption{Ablation study on Model Merging and Pass@k, reporting \textit{stability} (not benchmark performance) as measured by Kendall’s rank correlation ($\tau$). Merge@$N$ denotes averaging the last $N$ checkpoints. }
\label{tab:ablation_study}
\end{table}

\subsection{Ablation Study: Stabilizing Parameters with Checkpoint Merging}
Having established the effectiveness of the full framework, we now ablate its components, starting with Checkpoint Merging.

\paragraph{Smoothing Single-Run Trajectories. }

For tracking performance throughout a single training run,
% Figure \ref{fig:merge} visualizes the model’s performance trajectory across various benchmarks. We observe that evaluation scores of individual checkpoints exhibit significant volatility, at times resembling random fluctuations. In contrast, the merged model demonstrates a visibly smoother trend. 
Table \ref{tab:single_col_comparison} provides quantitative validation, reporting Kendall’s rank correlation ($\tau$) between evaluation scores and pre-training steps. Our merging approach achieves higher correlation on the majority of benchmarks, indicating a more stable measure of learning progress.

% \begin{figure}[h!]
%     \centering    \includegraphics[width=0.5\textwidth]{figs/merged_plots.pdf}
%   \caption{Evaluation stability visualization across different benchmarks.}
%   \label{fig:merge}
% \end{figure}

\begin{table}[h]
  \centering
  \small
    \begin{tabular}{@{} l c c r @{}}
      \toprule
      Benchmarks & Original & Merge@4 & \textbf{Improve} \\
      \midrule
      
      % --- 知识、考试与阅读理解分组 ---
      AGIEval       & 0.579& 0.845& \textbf{0.266}\\
      RACE          & 0.237& 0.745& \textbf{0.507}\\
      SQuAD2.0         & 0.042& 0.117& \textbf{0.075}\\
     \midrule
      
      MMLU          & 0.248& 0.200& \textbf{-0.048}\\
      CMMLU         & 0.705& 0.817& \textbf{0.111}\\
      CEval         & 0.533& 0.650& \textbf{0.117}\\
      \midrule
      
      % --- 数学能力分组 ---
      GSM8K         & 0.332& 0.343& \textbf{0.011}\\
      MATH          & 0.074& 0.393& \textbf{0.320}\\
      GSM-Plus      & 0.305& 0.498& \textbf{0.193}\\
      CMATH         & 0.220& 0.700& \textbf{0.481}\\
      \midrule
      
      % --- 代码能力分组 ---
      HumanEval     & 0.449& 0.570& \textbf{0.121}\\
      MBPP          & 0.216& 0.315& \textbf{0.099}\\
      MBPP-Plus     & 0.135& 0.198& \textbf{0.063}\\
      HumanEval-Plus & 0.426& 0.509& \textbf{0.083}\\
      
      \bottomrule
    \end{tabular}%
  % }
  \caption{Kendall's rank correlation ($\tau$) between evaluation scores and training progress under Original and Merge protocols.}
  \label{tab:single_col_comparison}
\end{table}

\paragraph{Reducing Inter-Run Variance.}

\begin{figure*}[h!]
    \centering
    \includegraphics[width=1\textwidth]{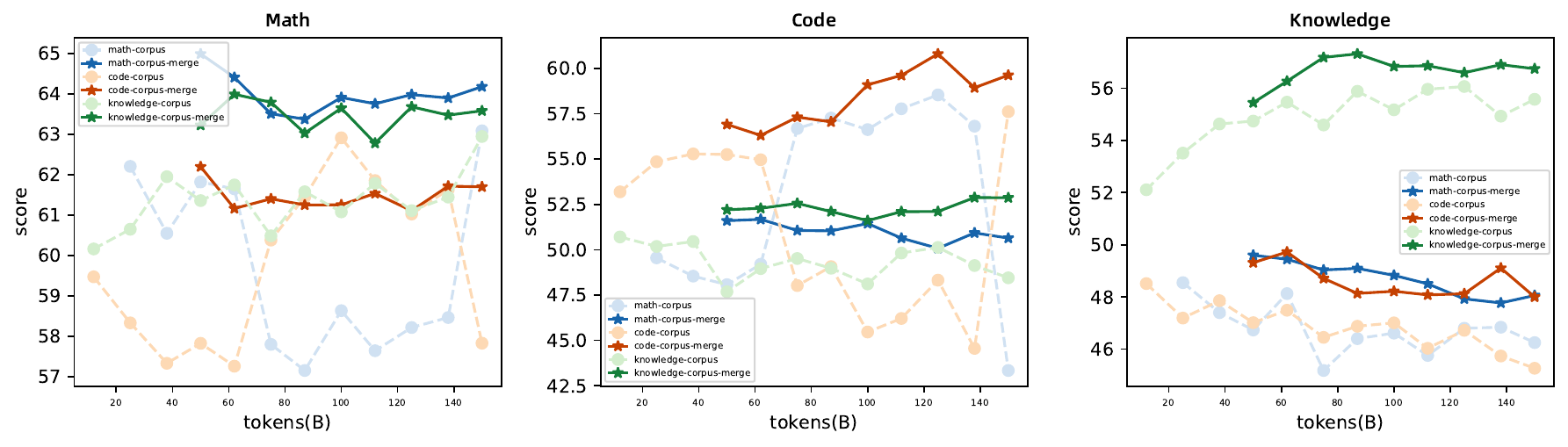}
    \caption{Checkpoint merging smooths training trajectories and clarifies model capabilities.}
    \label{fig:merge_2}
\end{figure*}
For ablation studies across multiple training runs, Figure~\ref{fig:merge_2} illustrates the performance of models trained on three distinct corpora (Math, Code, and Knowledge) on corresponding benchmarks, comparing individual checkpoints with their merged counterparts. The performance scores of original checkpoints fluctuate dramatically — exhibiting sharp drops and surges even in later training stages (e.g., for Math and Code), with trajectories frequently crossing. Judging solely from these volatile curves, it is difficult to determine which training strategy is truly optimal for a given capability. In contrast, the merged models eliminate this volatility. The resulting performance curves are significantly smoother and reveal a much clearer picture: the model trained on a specific corpus consistently and robustly outperforms others on its corresponding benchmark.

\paragraph{The Impact of $N$: Number of Merged Checkpoints.}
A key hyperparameter for merging is the number of checkpoints ($N$) included in the merging. Intuitively, a larger $N$ should increase stability by mitigating the impact of any single outlier checkpoint, effectively smoothing the parameter space. We empirically study this relationship by varying the merge window size. As shown in Table \ref{tab:ablation_merge}, when the checkpoint saving interval is fixed at 12.5B tokens, merging as few as 4 to 8 checkpoints yields substantial improvements in evaluation stability across most benchmarks.

\begin{table}[h]
  \centering
  \resizebox{\columnwidth}{!}{%
    \begin{tabular}{@{} l c c c c @{}}
      \toprule
      Benchmarks & Original & Merge@4 & Merge@8 & Merge@12 \\
      \midrule
      
      % --- 知识、考试与阅读理解分组 ---
      AGIEval       & 0.579& \textbf{0.845}& 0.813& 0.722\\
      RACE          & 0.237& \textbf{0.745}& 0.718&0.479\\
      SQuAD2.0         & 0.042& 0.117& 0.116&\textbf{0.667}\\
     \midrule
      
      MMLU          & 0.248& 0.200& 0.282&\textbf{0.389}\\
      CMMLU         & 0.705& 0.817& 0.846&\textbf{0.944}\\
      CEval         & 0.533& 0.650& \textbf{0.718}&0.667\\
      \midrule
      
      % --- 数学能力分组 ---
      GSM8K         & 0.332& 0.343& \textbf{0.581}&0.500\\
      MATH          & 0.074& 0.393& \textbf{0.744}&0.722\\
      GSM-Plus      & 0.305& 0.498& \textbf{0.641}&0.479\\
      CMATH         & 0.220& \textbf{0.700}& 0.684&0.254\\
      \midrule
      
      % --- 代码能力分组 ---
      HumanEval     & 0.449& 0.570& \textbf{0.790}&0.389\\
      MBPP          & 0.216& 0.315& 0.158&\textbf{0.435}\\
      MBPP-Plus     & 0.135& 0.198& \textbf{0.474}&0.087\\
      HumanEval-Plus & 0.426& 0.509& \textbf{0.614}&0.423\\
      
      \bottomrule
    \end{tabular}%
  }
  \caption{Impact of the number of merged checkpoints ($N$) on training stability, measured by $\tau$.}
  \label{tab:ablation_merge}
\end{table}

\subsection{Ablation Study: Stabilizing Measurements with Pass@k}
\label{passk}
% Next, we analyze the effect of Pass@k in isolation.

\paragraph{Improving Monotonicity of Training Signal.}
We first analyze the effect of Pass@k on stabilizing evaluation during training. Table~\ref{tab:kendall} reports Kendall’s rank correlation coefficient ($\tau$) across different benchmarks. For generative benchmarks (Math and Code), as $k$ in Pass@k increases, the Kendall's $\tau$ value consistently rises. This indicates that Pass@k provides a more monotonic and stable signal of training progress compared to the high volatility of greedy decoding. Conversely, as hypothesized in Section~\ref{sec:passk}, we observe a sharp decline in consistency for multiple-choice (MC) benchmarks (Knowledge). This empirical evidence confirms that the limited answer space of MC tasks amplifies noise via random guessing rather than measuring robust capability. Consequently, our MaP framework exclusively recommends and applies Pass@k for generative tasks, while reverting to standard protocols for MC tasks.

\begin{table}[h]
\centering

\sisetup{
  detect-weight,
  mode=text
}

\resizebox{\columnwidth}{!}{%
  \begin{tabular}{l S[table-format=1.3]
                   S[table-format=1.3]
                   S[table-format=1.3]
                   S[table-format=1.3]
                   S[table-format=1.3]
                   S[table-format=1.3]}
  \toprule
  \multirow{2}{*}{Protocol} & \multicolumn{2}{c}{Math (Gen)} & \multicolumn{2}{c}{Code (Gen)} & \multicolumn{2}{c}{Knowledge (MC)} \\
  \cmidrule(lr){2-3} \cmidrule(lr){4-5} \cmidrule(lr){6-7}
  & {GSM8K} & {MATH} & {HumanEval} & {MBPP} & {MMLU} & {CMMLU} \\
  \midrule
  Greedy   & 0.548 & 0.387 & 0.424 & 0.608 & \textbf{0.893} & \textbf{0.910} \\
  Pass@1   & 0.498 & 0.415 & 0.514 & 0.724 & 0.815          & 0.500 \\
  Pass@2   & 0.536 & 0.437 & 0.541 & 0.729 & 0.819          & 0.349 \\
  Pass@4   & 0.554 & 0.466 & 0.543 & 0.673 & 0.708          & 0.304 \\
  Pass@16   & \textbf{0.719} & \textbf{0.500} & \textbf{0.578} & \textbf{0.809} &     /      & / \\
  \bottomrule
  \end{tabular}%
} % <-- resizebox 结束
\caption{Kendall's rank correlation ($\tau$) between evaluation scores and training progress under different protocols. For generation tasks, we generate $n = 16$ samples per problem and evaluate the metric at $k \in \{1, 2, 4, 16\}$. For multiple-choice (MC) tasks, we generate $n = 4$ samples per problem and evaluate at $k \in \{1, 2, 4\}$.}
\label{tab:kendall}
\end{table}

\begin{figure*}[h!]
    \centering
    \includegraphics[width=1\textwidth]{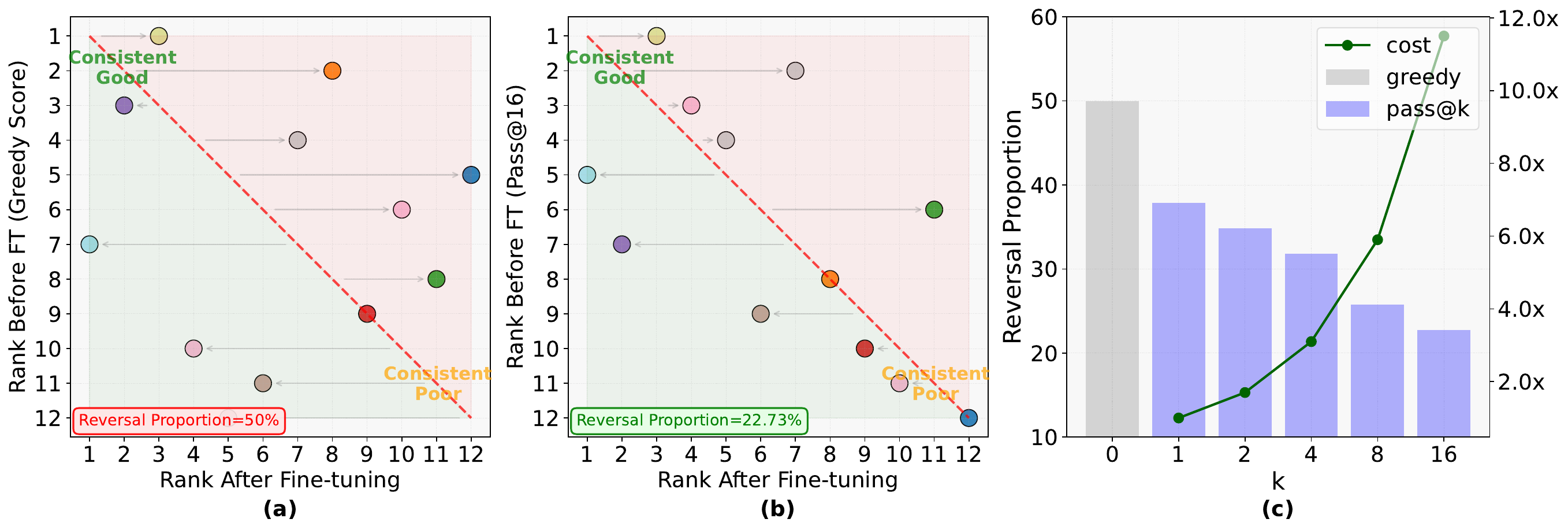}
    \caption{Pass@k improves the consistency between pre-training and post-SFT model rankings. (a) With greedy evaluation, the pre-training rank is a poor predictor of post-SFT rank, yielding a Pairwise Ranking Reversal Rate (PRR) of 50\%. (b) Using Pass@16 drastically improves consistency, reducing the PRR to 22.73\%. We generate $n=16$ samples per problem and calculate the metric for $k=\{1, 2, 4, 8, 16\}$. (c) The reversal proportion decreases monotonically as the sample count $k$ increases.}
    \label{fig:passk}
\end{figure*}

\paragraph{Enhancing Prediction of Downstream Performance.}
Next, we investigate whether this improved stability translates into better prediction of downstream performance. Figure~\ref{fig:passk} presents an experiment in which we train 12 smaller models (243M active parameters) using varied training strategies (e.g., learning rate schedulers) and then apply an identical SFT process to each. Figure~\ref{fig:passk}~(Left) plots pre-training ranks against post-SFT ranks using greedy evaluation. The Pairwise Ranking Reversal Rate (PRR) reaches 50\%, indicating that selecting a model for the next stage of the pipeline based on this evaluation is no better than random chance. In contrast, when using Pass@16 for evaluation (Figure~\ref{fig:passk}~(Middle)), the PRR drops significantly to 22.73\%. This trend is systematic, as shown in Figure~\ref{fig:passk}~(Right): the reversal proportion decreases monotonically as $k$ increases, confirming that a more stable evaluation protocol yields a more reliable forecast of downstream ranking.

% \subsection{Discussion}

\paragraph{The Impact of $k$: The Trade-off with Sample Budget.}
The choice of $k$ in Pass@k directly governs the trade-off between evaluation stability and computational cost. As demonstrated quantitatively in Table~\ref{tab:kendall} and visualized in Figure~\ref{fig:passk}~(Right), both stability and the reliability of model rankings improve monotonically as $k$ increases. However, this enhanced robustness comes at a significant cost. The line graph in Figure~\ref{fig:passk}~(Right) illustrates this near-linear growth in estimated evaluation cost with $k$ (see Appendix~\ref{cost} for details).
This establishes a clear need for researchers to balance their desired level of evaluation confidence with their available computational budget. To navigate this trade-off, future work could explore several promising directions. One practical strategy involves evaluating on smaller, carefully curated benchmark subsets that are highly representative of a model's capabilities. Another avenue is the development of dynamic sampling strategies, such as an early-stopping mechanism that terminates generation for a given problem once a correct solution is found.

\section{Related Work}

\subsection{LLM Evaluation and Benchmarking}
% The rapid advancement of large language models (LLMs) has been propelled by the development of comprehensive evaluation. Early efforts in natural language understanding, such as GLUE~\citep{glue} and SuperGLUE~\citep{superglue}, established a foundational paradigm for assessing model performance across diverse linguistic tasks. More recent benchmarks, including MMLU~\citep{mmlu}, MATH~\citep{math}, and HumanEval~\citep{humaneval}, have been designed to evaluate the broad, multi-task knowledge and reasoning capabilities of modern LLMs. While these benchmarks have played a pivotal role in defining what to evaluate, they often neglect the procedural question of how to obtain stable, reproducible measurements. Most public leaderboards report single-point estimates, implicitly disregarding the substantial instability we identify in this work. Our contribution complements this line of research not by introducing a new benchmark, but by proposing a more robust methodology for leveraging existing ones, thereby addressing the critical yet frequently overlooked issue of evaluation stability.
As models began to saturate these benchmarks, the community's focus shifted towards more challenging and holistic evaluations designed to probe the broad, multi-task knowledge and reasoning capabilities of modern LLMs.
Consequently, a new generation of benchmarks has emerged, targeting complex capabilities such as expert-level knowledge (e.g., MMLU~\citep{mmlu}), mathematical problem-solving (e.g., MATH~\citep{math}), and code generation (e.g., HumanEval~\citep{humaneval}). While this explosion in benchmark creation has been pivotal in defining what to evaluate, it has often overshadowed the critical procedural question of how to obtain stable, reproducible measurements. Most public leaderboards report single-point estimates, a practice that implicitly disregards the substantial performance variance we identify in this work and can lead to fragile or even misleading conclusions about model superiority. Our contribution complements this line of research not by introducing a new benchmark, but by proposing a more robust methodology for leveraging existing ones, thereby addressing the critical yet frequently overlooked issue of evaluation stability.

\subsection{Stability and Robustness of Evaluation}
The sensitivity of deep learning models, including LLMs, to stochastic factors such as weight initialization, data shuffling, and sampling randomness during inference is a well-documented phenomenon~\citep{DAmour2020,Bouthillier2021}. This inherent variability can lead to significant performance fluctuations across training or inference runs, even under identical hyperparameter configurations, making it challenging to distinguish genuine improvements from statistical noise. Moreover, \citet{ranjan2024post} introduces a phenomenon where performance trends observed in base machine learning models are inverted after applying post-hoc transforms. Prior work has quantified this variability through repeated experimental trials \citep{Bouthillier2021,philipp2018measuring}. In the context of LLMs, \citet{bose} improved evaluation stability by reformulating prompts and converting option-perplexity scoring into a fill-in-the-blank format. \citet{Instability} proposes integrating neighboring checkpoints through averaging or ensembling to mitigate performance fluctuations in LLM training.
In contrast to these approaches, our work provides a more systematic diagnosis of evaluation instability by identifying its two primary sources and offering a targeted, synergistic solution for each.
\section{Conclusion}
In this paper, we identify the critical problem of instability in Large Language Model evaluation. We decouple the issue into two primary sources: parameter instability from the stochasticity of training, and evaluation instability from fragile measurement protocols. To mitigate these, we propose a simple and effective dual-pronged framework combining Checkpoint Merging and the Pass@k metric. Our experiments demonstrate that this approach significantly smooths performance trajectories, reduces inter-run variance, and ensures more consistent model rankings. By providing a more reliable and reproducible evaluation paradigm, our work establishes a more solid foundation for the future development of large language models.
As future works, we plan to focus on enhancing the computational efficiency of this framework (e.g., adaptive sampling techniques), and a deeper investigation into the underlying causes of training volatility across different model scales and architectures will be crucial for developing inherently more stable training paradigms.

\section*{Limitations}
This work primarily addresses the symptoms of evaluation instability; a deeper inquiry into the fundamental causes of training volatility—be it optimization dynamics or data sequencing effects—remains a crucial avenue for future research that could lead to inherently more stable training paradigms. Moreover, the enhanced stability afforded by our framework represents a trade-off with computational resources. The Pass@k metric inherently requires multiple forward passes, and checkpoint merging necessitates the storage and processing of several checkpoints. 

% \section*{Acknowledgments}

% This document has been adapted
% by Steven Bethard, Ryan Cotterell and Rui Yan
% from the instructions for earlier ACL and NAACL proceedings, including those for
% ACL 2019 by Douwe Kiela and Ivan Vuli\'{c},
% NAACL 2019 by Stephanie Lukin and Alla Roskovskaya,
% ACL 2018 by Shay Cohen, Kevin Gimpel, and Wei Lu,
% NAACL 2018 by Margaret Mitchell and Stephanie Lukin,
% Bib\TeX{} suggestions for (NA)ACL 2017/2018 from Jason Eisner,
% ACL 2017 by Dan Gildea and Min-Yen Kan,
% NAACL 2017 by Margaret Mitchell,
% ACL 2012 by Maggie Li and Michael White,
% ACL 2010 by Jing-Shin Chang and Philipp Koehn,
% ACL 2008 by Johanna D. Moore, Simone Teufel, James Allan, and Sadaoki Furui,
% ACL 2005 by Hwee Tou Ng and Kemal Oflazer,
% ACL 2002 by Eugene Charniak and Dekang Lin,
% and earlier ACL and EACL formats written by several people, including
% John Chen, Henry S. Thompson and Donald Walker.
% Additional elements were taken from the formatting instructions of the \emph{International Joint Conference on Artificial Intelligence} and the \emph{Conference on Computer Vision and Pattern Recognition}.

% Bibliography entries for the entire Anthology, followed by custom entries
%\bibliography{custom,anthology-overleaf-1,anthology-overleaf-2}

% Custom bibliography entries only
\bibliography{custom}

\appendix

\section{Experimental Settings}
\label{setup}

\paragraph{Model architecture}
The core architectures of our experimental 8B MoE model are detailed in Table~\ref{table:Architectures}. 
The model is configured with 20 layers and a hidden dimension size of 2048. Except for the first layer, all FFNs layers are replaced with MoE layers. We adopt the GQA attention mechanism~\citep{gqa} and integrate Rotary Position Embedding (RoPE)~\citep{rope}, enabling the model to support sequence lengths up to 8K tokens. For parameter initialization, all learnable parameters are randomly initialized using a standard deviation of 0.006. 
Under this configuration, the model consists of a total of 8.6 billion parameters, of which approximately 1.43 billion are activated for each token during inference.

% \begin{table}[h]
%   % \small
%   \caption{{Detailed model architectures.}}
%   % \vspace{0.3cm}
%   \label{table:Architectures}
%   \centering
%   \resizebox{0.5\textwidth}{!}{
%   \begin{tabular}{cccccccccccc}
%     \toprule
%       & $n_{layers}$ & $d_{model}$ & $d_{ffn}$ & $d_{expert}$ &$n_{heads}$ & $n_{kv\_head}$ & $E$ & $E_a$ & $E_s$ & $N$ & $N_a$ \\
%     \midrule
%       & 20 & 2048 & 5120 & 512 & 16 & 4 & 256 & 8 & 1 & 16.3B & 1.43B \\
%     \bottomrule
%   \end{tabular}
%   }
% \end{table}

\begin{table}[h]
  \caption{Detailed model architecture.}
  \label{table:Architectures}
  \centering
  \begin{tabular}{lr}
    \toprule
    Parameter & Value \\
    \midrule
    Number of layers ($n_{layers}$) & 20 \\
    Model dimension ($d_{model}$) & 2,048 \\
    FFN dimension ($d_{ffn}$) & 5,120 \\
    Expert dimension ($d_{expert}$) & 512 \\
    Number of attention heads ($n_{heads}$) & 16 \\
    Number of KV heads ($n_{kv\_head}$) & 4 \\
    Total experts ($E$) & 32 \\
    Active experts ($E_a$) & 8 \\
    Shared experts ($E_s$) & 1 \\
    Total parameters ($N$) & 8.6B \\
    Active parameters ($N_a$) & 1.43B \\
    \bottomrule
  \end{tabular}
\end{table}

\paragraph{Training hyperparameters}
We use the AdamW optimizer~\citep{adamw} with hyperparameters set as follows: $\beta_1 = 0.9$, $\beta_2 = 0.95$, and weight decay of 0.1. Gradient clipping~\citep{clipping} norm is set to 1.0. According to the scaling laws for MoE optimal hyper-parameters, the maximum learning rates were set to $3.74e{-4}$. The batch size is set to 2048, and with a maximum sequence length of 8K, each training batch contains 16M tokens.

\section{Experimental Settings for PRR Analysis}
To investigate the Pairwise Ranking Reversal Rate (PRR), we conduct a controlled experiment involving the pre-training and subsequent supervised fine-tuning (SFT) of multiple model variants. For this study, we utilize smaller-scale models that share the same architecture as our primary model but have a total of 243M parameters. The base pre-training is conducted on 152B tokens with a learning rate of $1e{-3}$ and a global batch size of 512. To generate a diverse set of models for ranking, we introduced variations in the training strategy; specifically, during the final 10\% of the pre-training phase, each model variant was trained using a different learning rate schedule.
Following pre-training, each of the resulting checkpoints is subjected to an identical SFT process to ensure a fair comparison. This process involve fine-tuning each model on a dataset of 100k samples for 5 epochs. This setup allowe us to isolate the effect of the pre-training evaluation protocol on predicting post-SFT performance rankings.
% legend
\section{Cost Estimation for Pass@k}
\label{cost}

Evaluating generative models with Pass@k, while enhancing robustness, introduces computational overhead. We estimate the relative evaluation cost using the following formula:
\begin{equation*}
\small
\begin{split}
    \text{Cost} ={}& (\text{input\_cost\_cache\_not\_hit} \times \text{input\_tokens}) \\
    & + (\text{input\_cost\_cache\_hit} \times \text{input\_tokens} \times (n-1)) \\
    & + (\text{output\_cost} \times \text{output\_tokens} \times n)
\end{split}
\end{equation*}
Here, input\_cost\_cache\_not\_hit represents the cost for the first prompt generation (cache miss), while input\_cost\_cache\_hit applies to subsequent generations where the prompt can be reused (cache hit). output\_cost is the cost per generated token, and n is the total number of samples generated per problem. This estimation references typical API pricing models for input and output tokens, and prompt-to-generation token ratio.

\begin{table}[h]
\centering
\caption{Cost Parameters for Pass@k Estimation (per unit)}
\label{tab:cost_params}
\begin{tabular}{lc}
\toprule
\textbf{Parameter} & \textbf{Value} \\
\midrule
Input Cost (Cache Miss) & 1.25 \\
Input Cost (Cache Hit) & 0.125 \\
Output Cost & 10 \\
Input Tokens & 4 \\
Output Tokens & 1 \\
\bottomrule
\end{tabular}
\end{table}
% \section{Evaluation Protocol}
% Experiments are conducted based on OpenCompass~\citep{contributors2023opencompass}.  The specific settings for each benchmark are detailed in Table \ref{tab:eval_protocol}.
% \begin{table}[h!]
% \centering
% \caption{Evaluation settings for all benchmarks.}
% \label{tab:eval_protocol}
% \begin{tabular}{@{} lccc@ {}}
% \toprule
% \textbf{Benchmark} & &\textbf{Shots} & \textbf{Temperature} \\
% \midrule
% AGIEval && 5 & 0.0 \\
% RACE & ppl & 0 & - \\
% SQuAD2.0 &gen& 1 & 0.0 \\
% \midrule
% MMLU && 5 & 0.0 \\
% CMMLU && 5 & 0.0 \\
% CEval && 5 & 0.0 \\
% \midrule
% GSM8K && 8 & 0.8* \\
% MATH && 4 & 0.8* \\
% GSM-Plus && 8 & 0.8* \\
% CMATH && 5 & 0.8* \\
% \midrule
% HumanEval && 0 & 0.8* \\
% MBPP && 3 & 0.8* \\
% HumanEval-Plus && 0 & 0.8* \\
% MBPP-Plus && 3 & 0.8* \\
% \bottomrule
% \end{tabular}%
% \end{table}

% \section{Theoretical Analysis of Checkpoint Merging}
\section{Variance Reduction with Temporal Correlation}
\label{app:variance_analysis}
In Section~\ref{sec:method_merge}, we presented the variance reduction analysis under the independence assumption. Here, we derive the general case accounting for the temporal correlation inherent in optimization trajectories.

Let the parameter vector be modeled as $\theta_t = \theta_t^* + \epsilon_t$, where consecutive noise terms $\epsilon_i$ and $\epsilon_j$ have variance $\Sigma$ and a correlation coefficient $\rho_{ij} = \text{Corr}(\epsilon_i, \epsilon_j)$. The variance of the merged model $\hat{\theta}_T = \frac{1}{N}\sum_{i=0}^{N-1} \theta_{T-i}$ is determined by the variance of the averaged noise $\bar{\epsilon}$:

\begin{equation}
\begin{aligned}
    \text{Var}(\bar{\epsilon}) &= \text{Var}\left(\frac{1}{N} \sum_{i=0}^{N-1} \epsilon_{T-i}\right) \\
    &= \frac{1}{N^2} \left[ \sum_{i=0}^{N-1} \text{Var}(\epsilon_{T-i}) + \sum_{i \neq j} \text{Cov}(\epsilon_{T-i}, \epsilon_{T-j}) \right] \\
    &= \frac{1}{N^2} \left[ N\Sigma + \Sigma \sum_{i \neq j} \rho_{ij} \right] \\
    &= \frac{\Sigma}{N} \left[ 1 + \frac{1}{N} \sum_{i \neq j} \rho_{ij} \right].
\end{aligned}
\end{equation}

Let $\bar{\rho} = \frac{1}{N(N-1)} \sum_{i \neq j} \rho_{ij}$ denote the average pairwise correlation between noise terms in the merging window. The equation simplifies to:
\begin{equation}
    \text{Var}(\hat{\theta}_T) = \frac{\Sigma}{N} \left[ 1 + (N-1)\bar{\rho} \right].
\end{equation}

This derivation shows that if checkpoints are independent ($\bar{\rho} \approx 0$), we achieve the maximal variance reduction factor of $N$. Conversely, strong correlation diminishes this benefit.

In our experimental setup, consistent with standard LLM pretraining protocols, checkpoints are saved at sparse intervals (e.g., every few billion tokens). In stochastic optimization algorithms, temporal correlation of the update trajectory decays exponentially~\cite{Stochastic}. Given the large interval between checkpoints, the noise terms become effectively decorrelated.
Therefore, the independence assumption serves as a highly accurate approximation, and the variance reduction factor is close to the theoretical upper bound of $N$.

% \subsection{Mechanistic Analysis: Why Weight Merging Outperforms Score Smoothing}
% Our empirical results show that the performance trajectory of the merged model consistently lies above the original checkpoint scores. We attribute this to the geometric properties of the loss landscape and the similar effect of weight averaging with learning rate decay.

\section{Detailed Derivation of Pass@k Estimator and its Variance}
\label{sec:appendix_pass_at_k}
Let the probability of generating a correct solution for a given problem in a single, independent attempt be denoted by $p$. We can model this outcome as a Bernoulli trial:
\[
X \sim \mathrm{Bernoulli}(p)
\]
For evaluation, we generate $n$ independent candidate solutions. The outcomes are a set of i.i.d. random variables, $X_1, \dots, X_n$. The total number of successful solutions, $S = \sum_{i=1}^n X_i$, follows a Binomial distribution:
\[
S \sim \mathrm{Binomial}(n,p).
\]
The metric of Pass@k is the probability of obtaining at least one successful solution when drawing $k$ samples. This target parameter, which we denote as $q_k$, is given by:
\[
q_k = 1 - (1-p)^k.
\]

In a practical evaluation setting, we generate $n$ samples and observe a specific realization of the random variable $S$. From this count of correct solutions, we construct the standard unbiased estimator for Pass@k, denoted $\widehat{q}_{k,n}$, which is based on the hypergeometric probability of drawing $k$ failures from the set of $n$ samples, given that $n-S$ of them are failures:
\[
\widehat{q}_{k,n} = 1 - \frac{\binom{n-S}{k}}{\binom{n}{k}}, \quad \text{for } n \ge k.
\]

While the exact variance of $\widehat{q}_{k,n}$ has a complex closed-form expression, a highly accurate approximation for sufficiently large $n$ can be derived using the Delta method. This approach analyzes the variance of a simpler, asymptotically equivalent estimator. First, we use the sample mean, $\hat{p} = S/n$, to estimate the underlying success probability $p$.
\[
\operatorname{Var}(\hat{p}) = \operatorname{Var}\left(\frac{S}{n}\right) = \frac{np(1-p)}{n^2} = \frac{p(1-p)}{n}.
\]
The target parameter $q_k$ is a function of $p$, which we define as $h(p) = 1 - (1-p)^k$. The Delta method provides an approximation for the variance of a function of a random variable, $\operatorname{Var}(h(\hat{p})) \approx [h'(p)]^2 \operatorname{Var}(\hat{p})$, where $h'(p)$ is the first derivative. In this case, $h'(p) = k(1-p)^{k-1}$. Substituting these components into the formula yields the approximate variance of the Pass@k estimator:
\begin{align*}
\operatorname{Var}(\widehat{q}_{k,n}) &\approx \operatorname{Var}(h(\hat{p})) \\
&\approx [k(1-p)^{k-1}]^2 \cdot \frac{p(1-p)}{n} \\
&= \frac{k^2(1-p)^{2(k-1)}p(1-p)}{n}.
\end{align*}

To appreciate the stability gained, we compare this to the variance of a single-sample Bernoulli trial, $\operatorname{Var}(X) = p(1-p)$. The ratio of the variances is:
\[
\frac{\operatorname{Var}(\widehat{q}_{k,n})}{\operatorname{Var}(X)} \approx \frac{k^2(1-p)^{2(k-1)}}{n}.
\]
% This result reveals two primary sources of variance reduction: the $1/n$ factor from sampling and the exponential factor $(1-p)^{2(k-1)}$, which provides a powerful, often dominant, source of suppression. 
In conclusion, the Pass@k metric, by leveraging a multi-sample estimator, replaces a high-variance measurement with a statistically robust estimate whose variance is drastically lower, leading to more reliable and reproducible evaluation results.

\section{Statistical Significance and Confidence Intervals}
\label{sec:appendix_stats}

To show the significance of our findings, we perform a statistical analysis of the Kendall's $\tau$ correlation coefficients on 1B dense model. We compute the 95\% Confidence Intervals (CI) using bootstrap resampling ($n=1,000$ iterations) and calculate two-sided $p$-values against the null hypothesis of no correlation ($\tau = 0$).
As shown in Table \ref{tab:stats}, the MaP framework consistently yields higher $\tau$ values compared to the baseline. In datasets like MBPP and GSM8K, MaP narrows the confidence intervals, demonstrating increased stability. For challenging generative tasks like HumanEval, MaP shifts the $\tau$ center away from zero, achieving statistical significance ($p < 0.05$). This confirms that MaP recovers a true, monotonic learning signal from the noisy pre-training trajectory.

\begin{table}[h]
\centering
\caption{Statistical significance and confidence intervals for Kendall's $\tau$ correlation between training progress and performance scores.}
\label{tab:stats}
\resizebox{0.5\textwidth}{!}{
\begin{tabular}{llccc}
\hline
{Dataset} & {Protocol} & {Kendall's $\tau$} & {95\% CI} & {$p$-value}\\ \hline
\multirow{2}{*}{HumanEval} & Baseline & 0.133 & [-0.314, 0.568] & 0.506 \\
 & {MaP} & \textbf{0.492} & {[-0.000, 0.892]} & \textbf{0.009} \\ \hline
\multirow{2}{*}{MBPP} & Baseline & 0.761 & [0.545, 0.911] & <0.001 \\
 & {MaP} & \textbf{0.862} & \textbf{[0.626, 1.000]} & {<0.001} \\ \hline
\multirow{2}{*}{GSM8K} & Baseline & 0.816 & [0.555, 0.982] & <0.001 \\
 & {MaP} & \textbf{0.950} & \textbf{[0.820, 1.000]} & {<0.001} \\ \hline
\multirow{2}{*}{MATH} & Baseline & 0.833 & [0.313, 1.000] & <0.001 \\
 & {MaP} & \textbf{0.891} & \textbf{[0.539, 1.000]} & {<0.001} \\ \hline
\end{tabular}
}
\end{table}

\section{Discussion}
\subsection{Why Checkpoint Merging over Score Averaging?}
\begin{figure}[h!]
    \centering
    \includegraphics[width=0.5\textwidth]{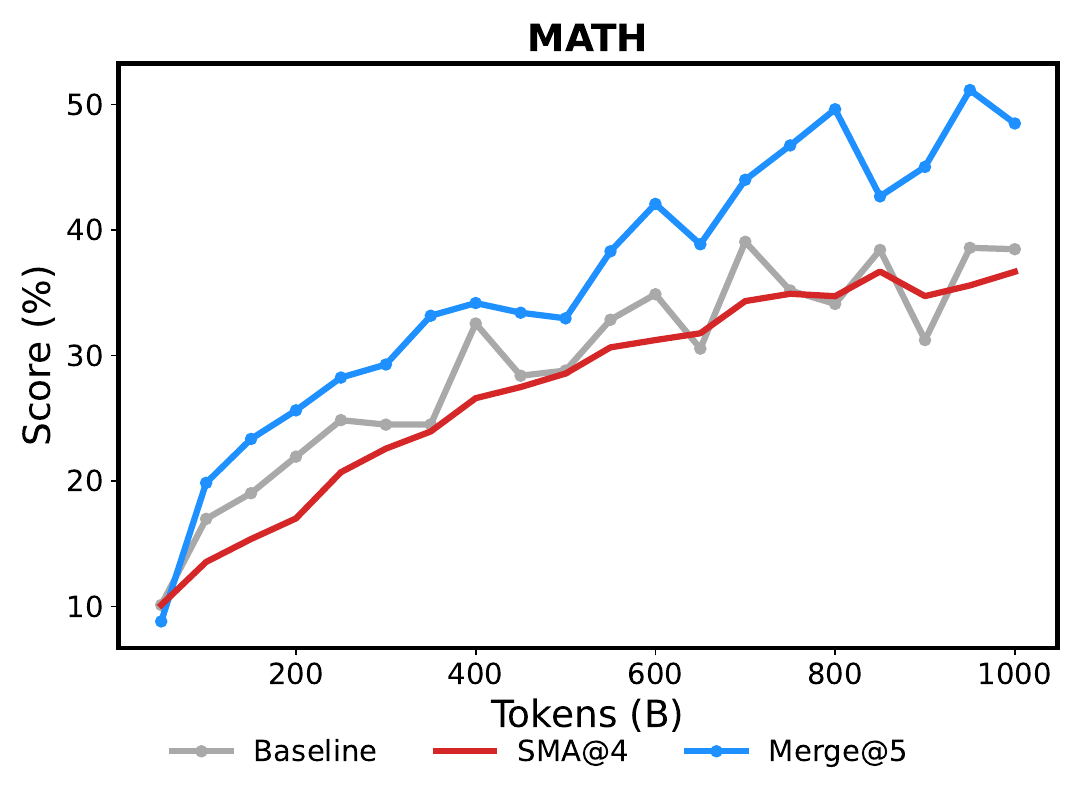}
    \caption{Visual comparison of Checkpoint Merging vs. Simple Moving Average (SMA). SMA applies a forced statistical smoothing to the scalar scores, which introduces a lagging effect and fails to reflect the model's actual capabilities. Conversely, Merge@5 denoises the high-variance transient states in the parameter space, accurately revealing the model's true and current learning dynamics}
    \label{fig:sma}
\end{figure}

A natural question arises: if the goal is to reduce evaluation variance, how does applying a Simple Moving Average (SMA) to the evaluation scores differ from Checkpoint Merging? While SMA can smooth the visualization of training dynamics, it fundamentally differs from our approach. 
SMA is merely a statistical average of historical scores. It yields a mediocre middle value within the window, rather than reflecting the model's true current capability. For instance, when the model is in a phase of rapid improvement, SMA is dragged down by lower scores from the past, thereby underestimating the model's current potential.
In contrast, MaP performs denoising in the parameter space, aiming to locate the centroid of the current training region. It represents the model's actual convergence state at the current moment, rather than just an average history. Consequently, the performance of MaP is higher than the average score of the $N$ points (the SMA value), and even surpass the single best checkpoint within the window, as illustrated in Figure~\ref{fig:sma}.

\subsection{What Are We Actually Measuring: Transient State vs. Latent Capability}
Another question regarding our approach is: if the merged model locates a better region in the parameter space, does evaluating it still reflect the effectiveness of the \textbf{current} checkpoint?
The goal of pretraining evaluation is not to measure a transient model state heavily influenced by recent data batches and stochastic noise (e.g., from dropout or sampling), but rather to assess the model’s underlying capability at its current training stage. Traditional single-point evaluation provides only a high-variance snapshot of this capability. In contrast, checkpoint merging mitigates the instantaneous bias introduced by training randomness, yielding a more faithful reflection of true progress, effectively approximating the performance attainable through strategies like learning rate annealing.

\end{document}